\pdfoutput=1

 \documentclass[final,3p,times,twocolumn]{elsarticle}

\usepackage{amssymb}
\usepackage{amsmath}

\usepackage{lineno}
\usepackage{graphicx}

\usepackage{booktabs}
\usepackage{pifont}
\usepackage{tabularx}
\usepackage{array}
\usepackage{multirow}
\usepackage{amsfonts}
\usepackage[linesnumbered,ruled,vlined]{algorithm2e}
\usepackage{xcolor}

\journal{Neural Networks}

\begin{document}

\begin{frontmatter}


\onecolumn 
\begin{highlights}
    \item A novel source-free domain adaptation method for cross-domain bearing fault diagnosis, balancing feature discriminability and diversity.
    \item A label voting strategy with data augmentation to classify and effectively utilize reliable and unreliable target samples.
    \item Superior performance compared to existing source-free methods and competitive results against non-source-free approaches.
\end{highlights}

\newpage 
\twocolumn 


\title{Source-free domain adaptation based on label reliability for cross-domain bearing fault diagnosis}


\author[A1]{Wenyi Wu}

\author[A2]{Hao Zhang}

\author[A3]{Zhisen Wei}

\author[A1]{Xiao-Yuan Jing}

\author[A1]{Qinghua Zhang}

\author[A1]{Songsong Wu\corref{mycorrespondingauthor}}

\cortext[mycorrespondingauthor]{Corresponding author}

\affiliation[A1]{organization={Guangdong Provincial Key Laboratory of Petrochemical Equipment Fault Diagnosis and School of Computer, Guangdong University of Petrochemical Technology},
            city={Maoming},
            postcode={525000},
            country={China}}

\affiliation[A2]{organization={Shenzhen Institute of Advanced Technology, Chinese Academy of Sciences},
            city={Shenzhen},
            postcode={518055},
            country={China}}

\affiliation[A3]{organization={Key Laboratory of Data Science and Intelligence Application (Minnan Normal University), Fujian Province},
            city={Zhangzhou},
            postcode={363000},
            country={China}}

\begin{abstract}
Source-free domain adaptation (SFDA) has been exploited for cross-domain bearing fault diagnosis without access to source data. Current methods select partial target samples with reliable pseudo-labels for model adaptation, which is sub-optimal due to the ignored target samples. We argue that every target sample can contribute to model adaptation, and accordingly propose in this paper a novel SFDA-based approach for bearing fault diagnosis that exploits both reliable and unreliable pseudo-labels. We develop a data-augmentation-based label voting strategy to divide the target samples into reliable and unreliable ones. We propose to explore the underlying relation between feature space and label space by using the reliable pseudo-labels as ground-truth labels, meanwhile, alleviating negative transfer by maximizing the entropy of the unreliable pseudo-labels. The proposed method achieves well-balance between discriminability and diversity by taking advantage of reliable and unreliable pseudo-labels. Extensive experiments are conducted on two bearing fault benchmarks, demonstrating that our approach achieves significant performance improvements against existing SFDA-based bearing fault diagnosis methods. Our code is available at https://github.com/BdLab405/SDALR
\end{abstract}


\end{frontmatter}




\section{Introduction}
\label{sec_Intro}

As critical components in rotating equipment, the performance and reliability of rolling bearings are vital for industrial machinery's stable operation and safety \cite{singh2020systematic}. Bearing faults can degrade equipment efficiency or lead to system failures, making fault diagnosis essential across industries such as manufacturing, petroleum refining, and aerospace \cite{ye2022multi,wu2023multiscale,cao2018mechanical,li2022dssdpp, shu2017detecting}.

Bearing fault diagnos is involves analyzing sensor signals (e.g., vibration, thermal images, acoustic emissions) to assess bearing health \cite{hoang2019survey, chen2022aligned}. While deep-learning-based methods offer improved accuracy, their reliance on the independence and identical distribution (i.i.d.) assumption between training and test samples limits real-world applicability. To address domain shifts, domain adaptation (UDA) methods transfer cross-domain knowledge for robust diagnostics \cite{zhuang2020comprehensive}. Metric-based approaches minimize domain discrepancies using measures like Maximum Mean Discrepancy (MMD) \cite{yan2017mind}, KL divergence \cite{nguyen2021kl}, and Wasserstein distance \cite{shen2018wasserstein}. Adversarial-based methods use Generative Adversarial Networks (GANs) to extract domain-invariant features \cite{benjdira2019unsupervised}.

\begin{figure}[htp]
  \centering
  \includegraphics[width=1\linewidth]{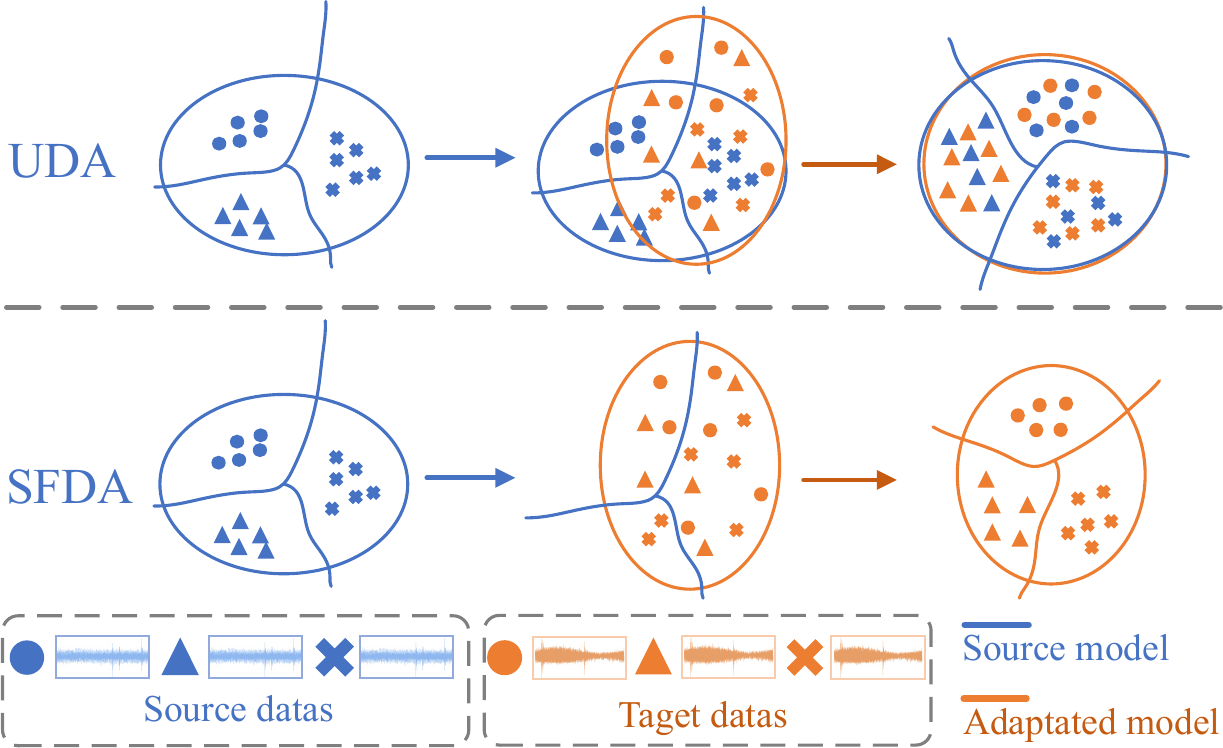}
  \caption{Domain Adaptation and Source Free Unsupervised Domain Adaptation.}
  \label{fig:sfda}
\end{figure}

Despite their effectiveness, UDA-based methods require access to source data, which is often impractical due to privacy concerns, storage, and computational burdens \cite{farahani2021brief, tang2019deep}. Source-Free Domain Adaptation (SFDA) addresses these limitations by adapting pre-trained source models to target domains without source data. Unlike conventional UDA methods, which rely on both source and target data to reduce domain discrepancies, SFDA enables compliance with data privacy regulations while reducing storage and computational demands, making it a practical solution for real-world applications (illustrated in Figure \ref{fig:sfda}).

Although different strategies such as self-training \cite{kim2021domain}, contrastive learning\cite{zhang2022divide}, clustering\cite{yang2021generalized} and data generation\cite{fang2024source} have been studied for SFDA.

Specifically, current SFDA-based bearing fault diagnosis approaches are prone to three nonnegligible obstacles\:(1) In bearing fault diagnosis, maintaining a balance between discriminability and diversity is essential for effective domain adaptation. SHOT\cite{liang2020we}, a foundational SFDA approach, emphasizes this principle and has been widely adopted by subsequent methods\cite{kumar2024mitigating, jianyu2023source}. However, many methods achieve a “fake balance” by relying solely on predicted target sample labels, ignoring the structural characteristics of target data and their role in label prediction. This limitation is particularly critical in fault diagnosis, where signal features are highly variable. (2) Vibration signals in bearing fault diagnosis are typically nonstationary, noisy, and influenced by operational conditions, exacerbating pseudo-label noise. Consequently, the reliability assessment of target pseudo-labels with the statistics of pseudo-labels, such as entropy\cite{zhu2023source}, can be inaccurate. Misclassified samples with high confidence often mislead the adaptation process, reducing the model's ability to accurately identify faults. (3) Distinguishing the importance of target samples is crucial in SFDA-based fault diagnosis. Due to the inherent correlations among vibration signals, unreliable samples often share features with reliable ones.  While existing methods\cite{zhu2023source, li2023source} prioritize reliable samples to improve adaptation, they tend to overlook the proper handling of unreliable samples. Inadequately processing these samples increases the risk of negative transfer, hindering the model's overall performance.

In this paper, we present a label-reliability-based source-free domain adaptation approach (SDALR) to address the above three issues: (1) We propose exploring the relation between feature space and label space to balance discriminability and diversity during model adaptation. The core principle is to ensure that the model produces similar features for samples sharing the same label and generate distinct features for samples with different labels. (2) To accurately assess the reliability of predicted labels in the target domain, we propose a pseudo-label voting strategy based on data augmentation. The strategy consists of separately predicting the class labels for each augmented sample and integrating the multiple pseudo-label by label voting. As a result, each target sample is given a specific label or is identified as an unreliable sample. (3) To alleviate negative transfer caused by the unreliable target samples, we propose to maximize the entropy of their predicted labels for less classification bias. In so doing, model adaptation is dominated by the increasing reliable samples and free from the inference of unreliable samples.
  
In summary, the contributions of this paper are as follows:
\begin{enumerate}
  \item We propose a novel source-free domain adaptation approach for cross-domain bearing fault diagnosis that explores the relation between vibration features and fault labels with well-balance between discriminability and diversity.
  \item We propose separate constraints on reliable and unreliable target samples which are obtained by our label voting strategy with data augmentation so that all target samples can be used for cross-domain adaptation.
  \item We empirically show the advantage of our approach over existent source-free methods for cross-domain bearing fault diagnosis and provide evidence for competitive performance against existing non-source-free works.
\end{enumerate}

The remainder of the paper is structured as follows. Section 2 introduces the related work. Section 3 describes details about the proposed method. Next, Section 4 reports the comprehensive experimental findings. At last, Section 5 concludes this study.

\section{Related Work}
\subsection{Diagnostic Methods Based on Unsupervised Domain Adaptation}
\label{sub2-1}
UDA methods generally require access to source data during the adaptation process. These approaches involve supervised learning on labeled source data while simultaneously measuring and reducing the distributional discrepancies between the source and target domains. Rooted in statistical learning theory, UDA techniques have been widely applied to various computer vision tasks \cite{cai2022dual}. By bridging these gaps, knowledge can be effectively transferred from the source domain to the target domain, enabling accurate predictions for unlabeled target data. Research in this area typically falls into two primary categories: metric-based methods and adversarial-based methods.

Metric-based methods, like MMD \cite{pan2010domain} and CORAL \cite{sun2016deep}, minimize distributional discrepancies through statistical measures. Wen et al. used MMD to extract domain-invariant features, while Qian et al. employed higher-order KL divergence for comprehensive moment alignment \cite{qian2019deep}. Ferracuti et al. applied Wasserstein distance for machine condition analysis \cite{ferracuti2021fault} and Chen et al. introduced UDAD to enhance knowledge transfer with mutual information maximization \cite{chen2020unsupervised}. Li et al. developed central moment metrics for precise fault diagnosis under varying conditions \cite{li2021central}.

Adversarial-based methods, inspired by GANs \cite{Tzeng_2017_CVPR}, automate feature extraction via adversarial training. Wang et al. integrated Wasserstein distance and domain classifiers in deep adversarial transfer learning \cite{wang2020intelligent}, while Kuang et al. combined self-supervised learning with dual classifiers \cite{kuang2022self}. Zhao et al. incorporated attention mechanisms into a multi-scale adversarial framework \cite{zhao2021deep} and Jiao et al. designed a cycle-consistent approach for domain alignment \cite{jiao2022cycle}.
Additionally, some studies have explored hybrid approaches that combine metric-based and adversarial-based strategies. For instance, Mao et al.'s multi-fault mode network \cite{mao2020new} and Zhou et al.'s adaptive transfer framework \cite{zhou2021deep}, blend metric- and adversarial-based strategies for dynamic alignment. Despite these advancements, traditional UDA's reliance on source data poses privacy challenges in sensitive contexts.

While traditional UDA methods have significantly advanced fault diagnosis under varying working conditions and distribution shifts, their reliance on source data presents notable limitations, particularly in scenarios where data privacy is a critical concern.

\subsection{Diagnostic Methods Based on Source-free Domain Adaptation}
\label{sub2-2}
In bearing fault diagnosis, SFDA focuses on adapting a pre-trained source model to an unlabeled target domain without requiring access to source domain data. This approach addresses the challenge of domain adaptation under restricted access to source data. SFDA is particularly significant in scenarios that demand stringent data privacy protections, as required by regulations such as the EU General Data Protection Regulation (GDPR) \cite{hoofnagle2019european} and China’s Data Security Law\cite{chen2021understanding}.

The concept of SFDA was first introduced by Liang \cite{liang2020we} in computer vision, SFDA fixes classifier parameters and generates high-confidence pseudo-labels for target data to self-train the feature generator. Building on this, Jiao \cite{jiao2022source} tailored SFDA for rotating machinery by combining label generation and nuclear norm regularization to preserve class diversity and enhance diagnostics. Similarly, Zhu \cite{zhu2022source} and Yue \cite{yue2022source} utilized information maximization to improve performance under varying conditions. Zhang \cite{zhang2023universal} expanded SFDA with a universal framework incorporating contrastive learning and self-supervised strategies for class alignment and unknown class rejection.

Clustering has also been investigated as an SFDA strategy, with its notable applications in computer vision serving as inspiration. Drawing from these works, Li \cite{li2023unsupervised} developed a convolutional neural network using fault mode clustering and attention mechanisms, while Zhu \cite{zhu2023source} and Zhang \cite{zhang2022source} introduced clustering-enhanced label generation approaches to boost diagnostic performance in target domains.

Despite these advancements, challenges remain in fully utilizing knowledge from the source model, generating high-quality pseudo-labels, and enhancing model robustness. The lack of direct reference information from the source model can lead to unstable training processes, potentially compromising model performance and reliability in real-world industrial applications.

\begin{figure*}[!t]
  \centering
  \includegraphics[width=0.9\linewidth]{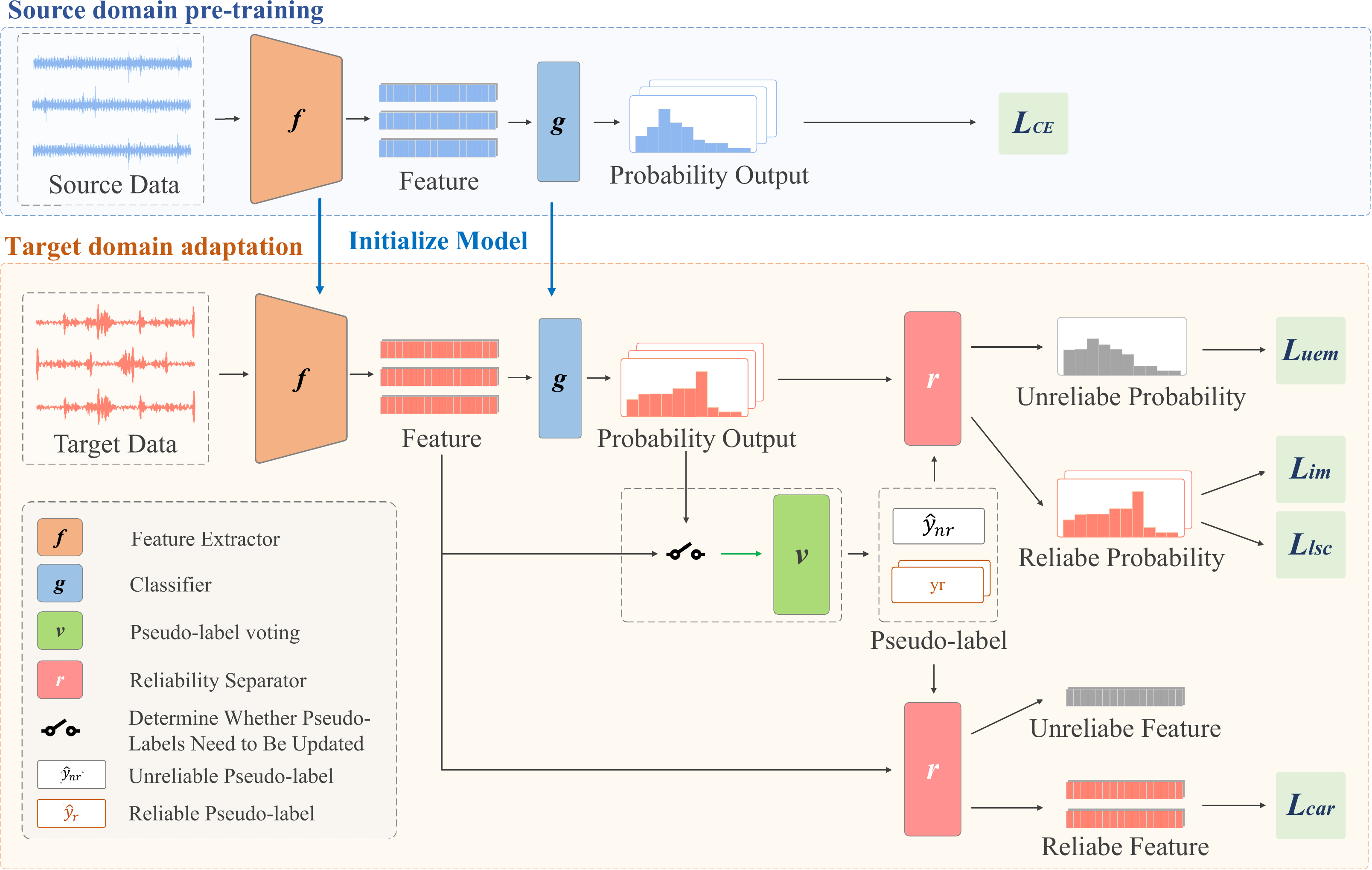}
  \caption{The method of SDALR. The source domain training phase is represented by the blue section, while the target domain adaptation phase is represented by the orange section.}
  \label{fig:method_overview}
\end{figure*}

\section{Proposed Method}

\subsection{Problem Definition}
This study addresses the cross-domain bearing fault diagnosis task as a $C$-way classification from a labeled source domain to a unlabeled target domain. We are given a source dataset $D_s=\{(x_i^s,y_i^s)\}_{i=1}^{N_s}$ from the source domain, where $x_i^s\in \mathcal{X}$ represents source data and $y_i^s \in \mathcal{Y}$ denotes its corresponding label. Meanwhile, we are also given a target dataset $D_t=\{x_i^t\}_{i=1}^{N_t}$, in which each target sample $x_i^t\in \mathcal{X}$ is sampled from the same data space as $D_s$ but its label $y_i^t \in \mathcal{Y}$ is unknown. In the context of domain adaptation, it is reasonably assumed that a marginal distribution shift exists between the source and target domains, i.e. $p_s(x) \ne p_t(x)$, while their posterio probability is identical, i.e. $p_s(y|x) \approx p_t(y|x)$. With respect to SFDA, a classification model $M_s$ is supervised trained on $D_s$, then it is adapted into the target domain merely relying on $D_t$ to predict the class labels $y_i^t$.

The source model is composed of a feature extractor $f_s$ and a classifier $g_s$, so that class probabilities are obtained by $p_s(x)=M_s(x)=\sigma(g_s(f_s(x)))$ for softmax function $\sigma$ and the predicted label $\hat{y}$ is provided by pseudo-label voting strategy. For adaptation to the target domain, a target model $M_t$ with the same network architecture as $M_s$ is initialized using the parameters of $M_s$ and then is updated on $D_t$ through knowledge transfer.

\subsection{Model Overview}
Figure \ref{fig:method_overview} presents the overall workflow of the SDALR method, which is composed of two major stages: the source domain pre-training stage and the target domain adaptation stage.

In the source domain pre-training stage, the source model is trained in a supervised manner using the labeled source domain dataset. Specifically, the model is optimized by minimizing the cross-entropy loss computed over the source domain dataset $D_s$.

After completing the source domain pre-training, the parameters of the source model $M_s$ are used to initialize the target model $M_t$ and the process transitions to the target domain adaptation phase. During this phase, the target model is fine-tuned using the unlabeled target domain dataset. For pseudo-label generation, a voting strategy based on data augmentation is used, categorizing samples into reliable and unreliable groups. Reliable samples are assigned their corresponding pseudo-labels, while unreliable samples are labeled as -1 to denote their unreliable. For reliable samples, the model is optimized using the cohesion and repulsion loss $L_{car}$, this loss function promotes the clustering of similar features and the segregation of dissimilar features. Inspired by the SHOT\cite{liang2020we} method, a combination of label-smoothing cross-entropy loss $L_{lsc}$ and information maximization loss $L_{im}$ is used to further refine the model's alignment with reliable samples. For unreliable samples, the entropy maximization loss $L_{uem}$ is applied, which maximizes entropy to limit the certainty of unreliable samples.

\subsection{Source Domain Training}

In the stage of source domain training, the objective of source domain training is to develop a well-generalized source model $M_s$, which, after being trained on $D_s$, serves as the initialization for the target model.


During the training of the source model, the classic cross-entropy loss is used as the optimization objective. The optimization objective can be formulated as follows:

\begin{equation}\label{Eq:Lce}
    \small
    L_{CE}=-\mathbb{E}_{x_i^s\in D_s}\sum_{c=1}^C q_c \log p^c(x_i^s)
\end{equation}
where C is the number of classes. $q$ is the onehot encoding of $y_i^s$ so that $q_c$ is '1' for the correct class and '0' for the rest over class c and $p^c(x_i^s)$ denotes the predicted probability of class $c$ for the sample $x_i^s$, as generated by the source model, also can expressed as $p^{c}(x_{i}^{s}) = \sigma^{c}(g(f(x_{i}^{s})))$.


\subsection{Target Domain Adaptation}
\begin{figure*}[!tph]
  \centering
  \includegraphics[width=0.9\linewidth]{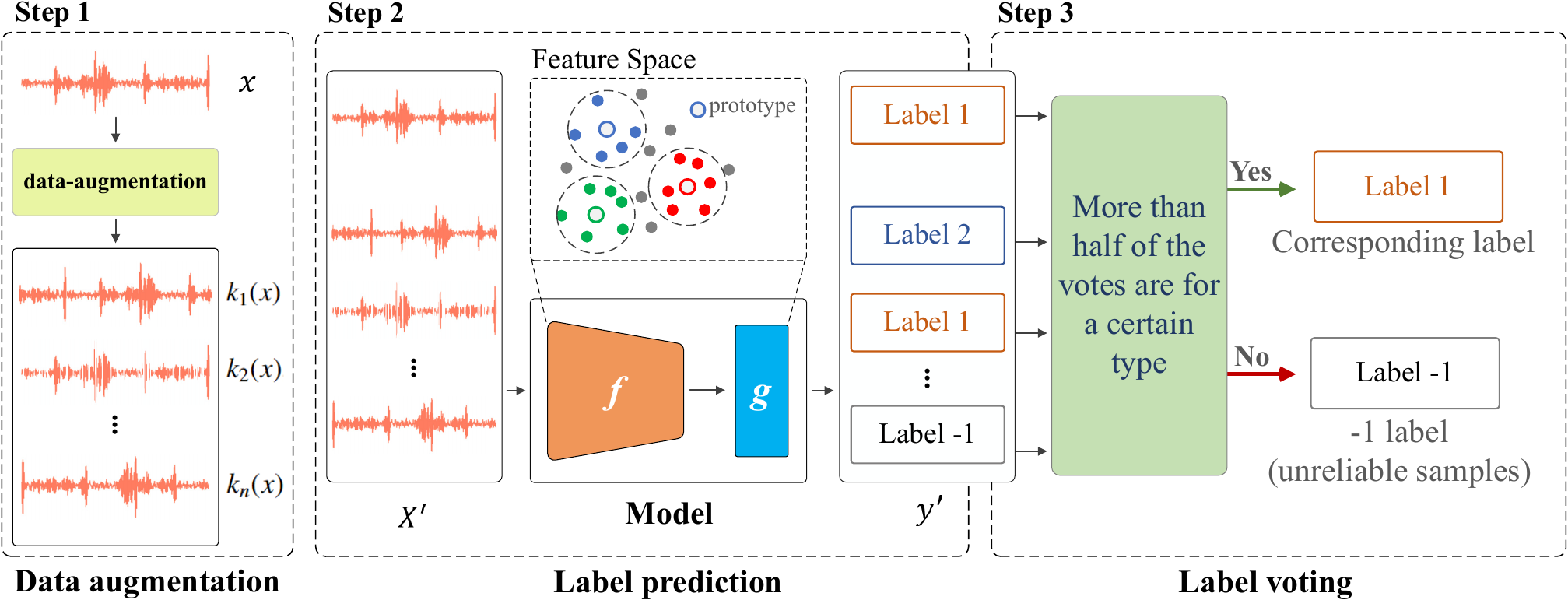}
  \caption{The specific process of pseudo-label voting}
  \label{fig:voting}
\end{figure*}
In target domain adaptation stage, the target model $M_t$ is initialized with the parameters of the pre-trained source model $M_s$. Subsequently, the adaptation process is performed using the unlabeled target domain dataset. To address this, the study introduces the SDALR approach for effective target domain adaptation.

\subsubsection{Pseudo-Label Generation}
While some methods assess the reliability of pseudo-labels using statistical metrics, they are often prone to inaccuracies due to noise in the pseudo-labels. To address this, this study proposes an innovative approach. The specific process is shown in Figure \ref{fig:voting}.

\textbf{Data augmentation.}   
each sample undergoes n distinct data augmentation transformations, resulting in an expanded set of augmented instances, represented as:
\begin{equation}\label{Eq:X'}
    \small
    X'=\{x,k_1(x),k_2(x),\ldots,k_n(x)\}
\end{equation}
where $k_1(x),k_2(x),\ldots,k_n(x)$ denote n types of data augmentation methods. In this study, we apply three specific augmentation techniques: flipping, random zeroing, and cyclic shifting. Consequently, the set is given by: $X^{'} =\{x,k_1(x),k_2(x),k_3(x)\}$, where $k_1(x)$, $k_2(x)$, and $k_3(x)$ correspond to flipping, random zeroing, and cyclic shifting, respectively.

\textbf{Label prediction.}  
We adopt a strategy combining prototype-based (class-wise feature center) calculations with confidence filtering, aiming to generate more reliable pseudo-labels.

Specifically, the initial prototype for each class is calculated by integrating the predicted probabilities (Softmax outputs) and the corresponding feature vectors of target domain samples. The prototype $\mathrm{n}_c$ for class $c$ is computed as follows:
\begin{equation}\label{Eq:nc}
    \small
    \eta_c=\frac{\sum_{i=1}^{N_t}p^c(x_i^t)f(x_i^t)}{\sum_{i=1}^{N_t}p^c(x_i^t)}
\end{equation}
where $f(\cdot)$ represents the feature vector of a sample, $p^{c}(\cdot)$ denotes the predicted probability for class $c$ and $p^c(x_i^t)$ probability of sample $x_i^t$ belonging to class $c$.

With the class-wise prototypes computed, initial pseudo-labels are assigned based on a similarity threshold. The similarity metric measures the alignment of a sample's feature vector with its corresponding class prototype. A sample's initial pseudo-label $\hat{y}^0(x)$ determined as:
\begin{equation}\label{Eq:sc}
    \small
    s_c(x)=\frac{f(x)\cdot\eta_c}{\parallel f(x)\parallel\parallel\eta_c\parallel}
\end{equation}
\begin{equation}\label{Eq:y0}
    \small
    \hat{y}^{0}(x)=\begin{cases}\arg\max_cs_c(x),&\text{if}\max_cs_c(x)>\partial\\-1,&\text{otherwise}\end{cases}
\end{equation}
where $s_c(x)$ resents the cosine similarity between the feature vector $f(x)$ and the prototype $\mathrm{n}_c$, with $\left\|\cdot\right\|$ denoting the Euclidean norm. The threshold $\partial$ determines whether a pseudo-label is assigned; samples below this threshold are labeled as -1, indicating an unreliable sample.

\textbf{Label voting. }
Using the formula above, we compute the label set for each sample in the augmented set $X'$. The label set is expressed as:
\begin{equation}\label{Eq:Y'}
    \hat{Y}'=\{\hat{y}_0^0(x),\hat{y}_1^0(x),\hat{y}_2^0(x),\ldots,\hat{y}_n^0(x)\}
\end{equation}
where $\hat{y}_n^0(x)$ is the pseudo-label for the augmented instance $k_n(x)$ and $\hat{y}_0^0(x)$  corresponds to the original sample $x$. These labels are aggregated using a majority voting strategy. The final pseudo-label $\hat{y}$ is determined as:
\begin{equation}\label{Eq:y}
    \small
    \hat{y}=
    \begin{cases} 
        \mathrm{argmax}_k\left(\sum_{j=1}^m \mathbb{I}(\hat{y}_{j}^{0}=k)\right), & 
        \mathrm{if} \mathrm{max}\left(\sum_{j=1}^m \mathbb{I}(\hat{y}_{j}^{0}=k)\right) > \frac{m}{2} \\
        -1, & \mathrm{otherwise}
    \end{cases}
\end{equation}
where m denotes the number of pseudo-labels in the voting process and $\mathbb{I}(\cdot)$ is an indicator function that outputs 1 if the condition is true and 0 otherwise. According to this formula, a pseudo-label is finalized if it receives more than half the votes; otherwise, the sample is assigned -1.

To mitigate the adverse effects of imbalanced pseudo-label distributions, this study leverages data augmentation techniques to generate additional samples, thereby achieving a balanced pseudo-label distribution. Specifically, for classes with fewer samples, excluding the label -1,  a subset of samples is randomly duplicated and each duplicated sample undergoes one of several randomly selected augmentation methods, including flipping, random masking, or circular shifting. This approach ensures that the number of samples in each class is balanced. 

\begin{figure*}[!th]
  \centering
  \includegraphics[width=0.7\linewidth]{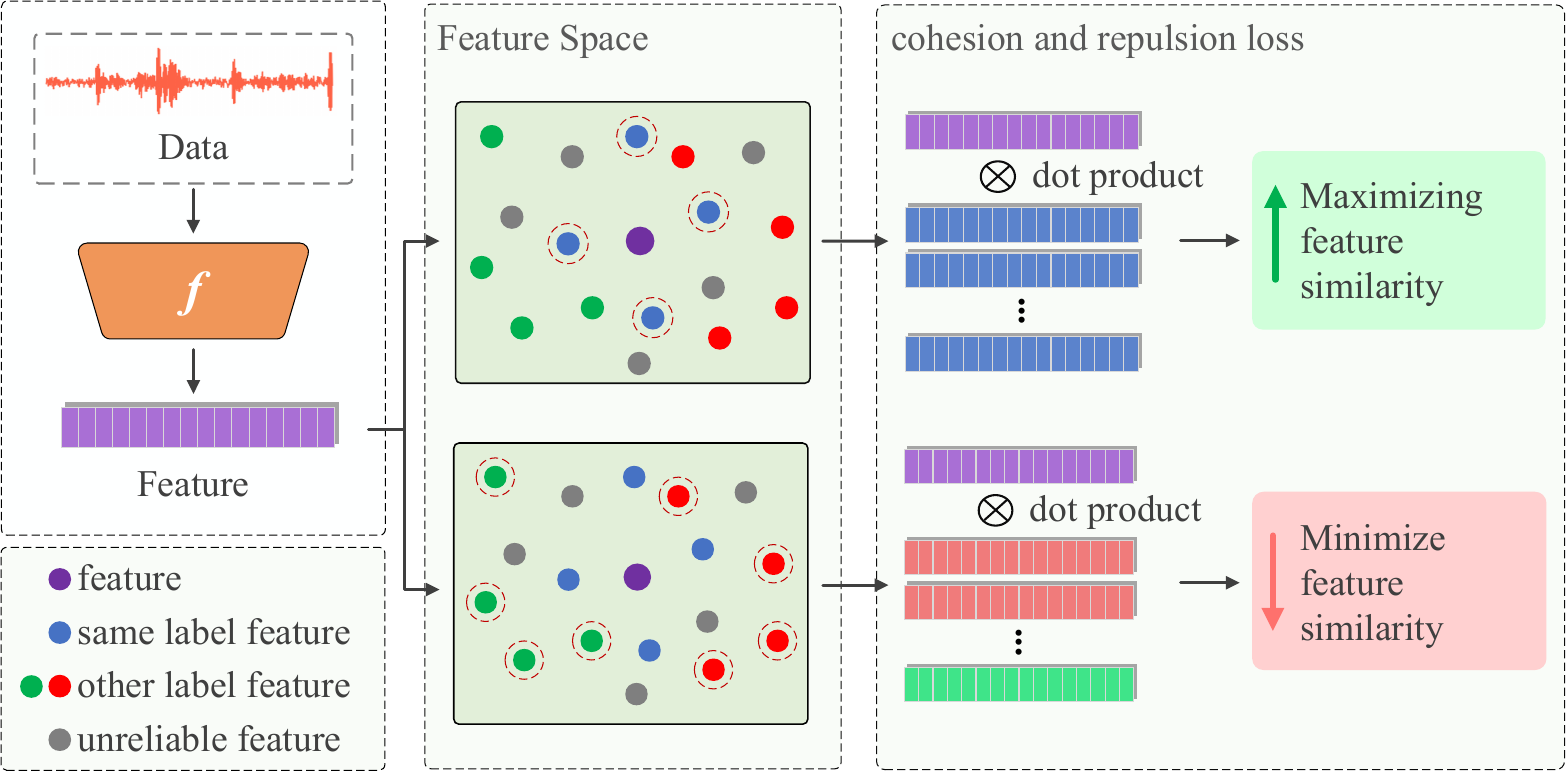}
  \caption{The specific process of cohesion and repulsion}
  \label{fig:add}
\end{figure*}

\subsubsection{Label-Smoothing Cross-Entropy Optimization}

To address the noise and uncertainty introduced during pseudo-label generation, we use a self-training approach with a label-smoothing cross-entropy loss $L_{lsc}$. The objective function for this process is defined as follows:
\begin{equation}\label{Eq:L_lsc}
    \small
    L_{lsc}=-\mathbb{E}_{x_{i}^{t}\in D_{t}}\sum_{c=1}^{C}\mathbb{I}(\hat{y}_{i}\neq-1)q_{(c,i)}^{ls}\log p^{c}\left(x_{i}^{t}\right)
\end{equation}
where the indicator function $\mathbb{I}(\hat{y}_i\neq-1)$ acts as a filtering mechanism, excluding samples labeled as -1 from the loss computation. $p^{c}(x_{i}^{t})$ is the model's predicted probability for class $c$. $q_{(c,i)}^{ls}$ represents the smoothed pseudo-label distribution for sample $x_i^t$ in class $c$, which is defined as:
\begin{equation}\label{Eq:q}
    \small
    q_{(c,i)}^{ls}=(1-\alpha)q_{(c,i)}+\frac{\alpha}{C}
\end{equation}
where $q_{(c,i)}$ is the one-hot encoding of the pseudo-label $\hat{y}_i$ assigned to sample $x_i^t$, $\alpha$ is a coefficient controlling the degree of smoothing and $\alpha$ is generally defined as 0.1.

\subsubsection{Unreliable Samples Entropy Maximization}
Directly minimizing entropy for Unreliable samples can exacerbate errors, while ignoring them may bias the model toward reliable samples. To address this, we apply entropy maximization for unreliable samples, reducing the model's overconfidence in their predictions. This is captured in the entropy maximization loss $L_{uem}$:
\begin{equation}\label{Eq:L_uem}
    \small
    L_{uem}=\mathbb{E}_{x_i^t\in D_t}\left[\mathbb{I}(\hat{y}=-1)\sum_{c=1}^Cp^c(x_i^t)\log p^c(x_i^t)\right]
\end{equation}
where $p^c(x_i^t)$ represents the predicted probability of target sample $x_i^t$ for class $c$. $L_{uem}$ calculates the entropy for samples where $\hat{y}=-1$, minimizing it to enhance the model's confidence in reliable samples. 

\subsubsection{Information Maximization}
To improve model reliability and achieve balanced class distributions, we draw inspiration from the SHOT\cite{liang2020we} framework in handling reliable samples. Specifically, we use an information maximization loss comprising two sub-losses designed. The first sub-loss, $L_{ent}$, minimizes the prediction entropy for samples with reliable pseudo-labels ($\hat{y}\neq-1$). This encourages the model to make precise and confident decisions for these samples. The second sub-loss, $L_{div}$, promotes balanced class distributions by penalizing deviations from a uniform distribution across all classes. The mathematical formulations for these sub-losses are as follows:
\begin{equation}\label{Eq:L_ent}
    \small
    L_{ent}=-\mathbb{E}_{x_i^t\in D_t}\left[\mathbb{I}(\hat{y}\neq-1)\sum_{c=1}^Cp^c(x_i^t)\log p^c(x_i^t)\right]
\end{equation}
\begin{equation}\label{Eq:L_div}
    \small
    L_{\mathrm{div}}=\mathbb{I}(\hat{y}\neq-1)\sum_{c=1}^C\hat{p}^c\log\hat{p}^c
\end{equation}
where $\hat{p}^c$ denotes the mean predicted probability for class c across the target dataset, defined as $\hat{p}^{c} =\mathbb{E}_{x_i^t\in D_{t}}[\sigma^{c}(g(f(x_i^t)))]$.

The IM loss integrates entropy-based constraints and diversity regularization, formulated as:
\begin{equation}\label{Eq:L_im}
    \small
    L_{\mathrm{im}}=L_{ent}+L_{\mathrm{div}}
\end{equation}

\subsubsection{Optimization for Feature Cohesion and Repulsion}
During target domain adaptation stage, it is crucial to maintain a balance between feature discriminability and diversity. Neglecting the intrinsic data structure of the target domain and its influence on label predictions limits the model’s capacity to effectively balance these two aspects. To address this limitation, we propose the cohesion and repulsion loss $L_{car}$. The specific process is shown in Figure \ref{fig:add}.

For each target sample $x_i^t$ with reliable predicted pseudo-label $\hat{y}\neq-1$, we obtain two sample sets as follows,
\begin{enumerate}
	\item \textbf{Similar set $S_i$} includes reliable target samples that share the same pseudo-label as $x_i^t$. In the practical phase of training, $S_i$ is constructed within each mini-batch.
	\item \textbf{Background set $N_i$} comprises reliable target samples that do not belong to $S_i$. In the practical phase of training, $N_i$ consists of reliable samples in each mini-batch with different pseudo-labels of $x_i^t$.
\end{enumerate}

We define the cohesion and repulsion loss function as:
\begin{equation}\label{Eq:L_car}
    \small
    L_{car}=\mathbb{E}_{x_i^t\in X_t}\left[\sum_{x_j^t\in S_i}f(x_i^t)^Tf(x_j^t)-\beta\sum_{x_m^t\in N_i}f(x_i^t)^Tf(x_m^t)\right]
\end{equation}
where $f(x_i^t)^T$ represents the transpose of $f(x_i^t)$ and $f(x_i^t)^Tf(x_j^t)$ denotes their inner product, measuring similarity. The goal is to maximize the similarity between $x_i^t$ and samples in $S_i$ by maximizing $\sum_{x_j^t\in S_i}f(x_i^t)^Tf(x_j^t)$ while simultaneously minimizing the similarity between $x_i^t$ and samples in $N_i$ by minimizing $\sum_{x_m^t\in N_i}f(x_i^t)^Tf(x_m^t)$. To mitigate the influence of background features, the parameter $\beta$ is fixed at a constant value of 0.6.

\subsubsection{The overall loss function}
In the end, \ref{Eq:L_lsc}, \ref{Eq:L_uem}, \ref{Eq:L_im} and \ref{Eq:L_car} are conbined for Target Domain Adaptation. The integrated training objective is expressed as follows:
\begin{equation}\label{Eq:final_loss}
    \small
    L_{tar}=L_{lsc}+L_{uem}+L_{im}+L_{car}
\end{equation}

The whole process of the proposed method SDALR is shown in Algorithm \ref{algorithm 1}.

\begin{algorithm}[!thp]
\caption{The process of SDALR}\label{algorithm 1}
\small
\SetKwInput{KwInput}{Input}
\SetKwInput{KwOutput}{Output}
\KwInput{Target domain dataset $D_t=\{x_i^t\}_{i=1}^{N_t}$, threshold $\partial$, parameter $\beta$, source domain model $M_s$}
\KwOutput{Transferred target model $M_t$}

\For{each training iteration}{
    \If{pseudo-labels need updating}{
        Augment $D_t$ to create enhanced sample set $X^{\prime}$ (Eq.\ref{Eq:X'})\;
        Extract features $f(x)$ from $D_t$ using $f$ and calculate centroids $\eta_{c}$ (Eq.\ref{Eq:nc})\;
        Compute similarity $s_c$ between $X^{\prime}$ and $\eta_{c}$ (Eq.\ref{Eq:sc})\;
        Filter outputs $\hat y^{0}$ using threshold $\partial$ (Eq.\ref{Eq:y0})\;
        Update pseudo-labels $\hat y_i$ through voting strategy (Eq.\ref{Eq:y})\;
        Generate additional samples for each class, with each additional sample randomly chosen a data augmentation method\;
    }
    Extract features $g(x_i^t)$ and probability outputs $p(x_i^t)$ for target samples $x_i^t$ using $M$\;
    \If{label of $x_i^t$ is -1}{
        Calculate loss $L_{uem}$ (Eq.\ref{Eq:L_uem})\;
    }\Else{
        Calculate loss $L_{lsc}$ (Eq.\ref{Eq:L_lsc})\;
        Calculate loss $L_{ent}$ (Eq.\ref{Eq:L_ent})\;
        Calculate global entropy $L_{div}$ (Eq.\ref{Eq:L_div})\;
        Compute $L_{im}$ by combining $L_{ent}$ and $L_{div}$ (Eq.\ref{Eq:L_im})\;
        Retrieve all samples with the same label from the current batch as set $S_i$\;
        Retrieve all samples with different labels in the current batch as set $N_i$\;
        Calculate $L_{car}$ using $N_i$ and $S_i$ (Eq.\ref{Eq:L_car})\;
    }
    Combine $L_{lsc}$, $L_{im}$, $L_{uem}$ and $L_{car}$ to compute total loss (Eq.\ref{Eq:final_loss})\;
    Update model $M$ based on total loss\;
}
\end{algorithm}

\section{Experiments}
\label{sec:exp}

This section provides a comprehensive evaluation of the proposed SDALR method using two public datasets: the PU dataset and the JNU dataset. The experiments aim to demonstrate the superiority of SDALR in fault diagnosis tasks compared to existing methods and to analyze its effectiveness under various combinations of loss functions.

\subsection{Datasets}
\begin{figure}[htp]
  \centering
  \includegraphics[width=1\linewidth]{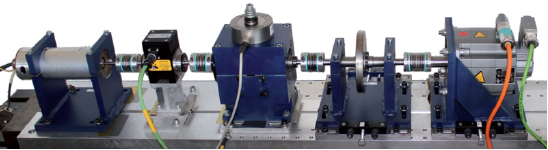}
  \caption{PU\cite{lessmeier2016condition} experimental platform demonstration}
  \label{fig:PU}
\end{figure}

\textbf{Paderborn University (PU) Dataset \cite{lessmeier2016condition}}: The PU dataset is a widely used benchmark in bearing fault diagnosis, providing diverse and complex data across various rolling bearing conditions. It consists of six healthy bearing samples, 12 artificially damaged samples and 14 naturally worn samples generated under accelerated loading. Data was collected at a sampling rate of 64 kHz, with 20 independent measurements per condition, each lasting four seconds. Figure \ref{fig:PU} illustrates the data collection platform. This study targets eight bearing types for classification: one healthy bearing (K001) and seven naturally faulted bearings (KA04, KA15, KA22, KA30, KI14, KI17 and KI21). The data length is truncated to 2048, with 2000 samples extracted per class. To evaluate performance under different working conditions, six learning tasks were created based on three domain settings: (1) Rotational speed of 1500 r/min, load torque of 0.1 Nm and radial force of 1000 N; (2) Rotational speed of 1500 r/min, load torque of 0.7 Nm and radial force of 400 N; (3) Rotational speed of 1500 r/min, load torque of 0.7 Nm and radial force of 1000 N. As shown in the Table \ref{tab:PU}, these configurations are referred to as domains $A_1$, $A_2$, and $A_3$, respectively.

\begin{table}[htbp]
  \centering
  \caption{The Domain Divisions of Datasets PU}
    \begin{tabularx}{0.45\textwidth}{>{\centering\arraybackslash}X>{\centering\arraybackslash}X}
    \toprule
    Domain & Parameter combination \\
    \midrule
    $A_1$    & N15\_M01\_F10  \\
    $A_2$    & N15\_M07\_F04  \\
    $A_3$    & N15\_M07\_F10  \\
    \bottomrule
    \end{tabularx}%
  \label{tab:PU}%
\end{table}%

\begin{figure}[htp]
  \centering
  \includegraphics[width=0.38\textwidth]{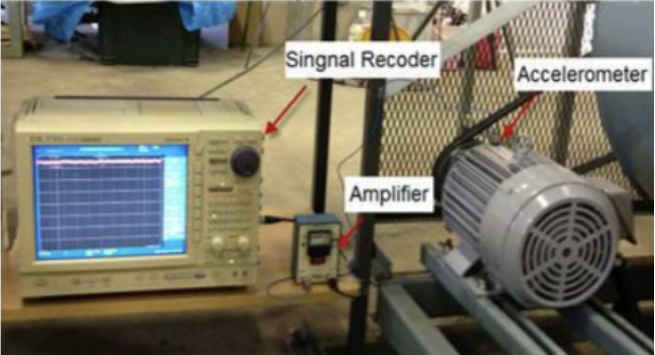}
  \caption{JNU\cite{li2013sequential} experimental platform demonstration}
  \label{fig:JNU}
\end{figure}

\textbf{Jiangnan University (JNU) Dataset \cite{li2013sequential}}: The JNU dataset, developed by Jiangnan University in China, is an essential resource for bearing fault diagnosis research. It includes four primary bearing types: healthy bearings, bearings with inner ring damage, outer ring damage and rolling element damage, labeled as H, IR, OR and B, respectively. The data length is truncated to 2048, with 2000 samples extracted per class. Faults were introduced through precision machining, creating micro-indents of 0.3 mm × 0.05 mm (width × depth) on the respective components. Vibration signals were captured using PCB MA352A60 accelerometers at a 50 kHz sampling rate, recording vertical vibrations. The data collection platform is illustrated in Figure \ref{fig:JNU}. As shown in the Table \ref{tab:JNU}, the dataset covers three rotational speed conditions, designated as domains $B_1$, $B_2$, and $B_3$ and supports the construction of six learning tasks.

\begin{table}[htbp]
  \centering
  \caption{The Domain Divisions of Datasets JNU}
    \begin{tabularx}{0.45\textwidth}{>{\centering\arraybackslash}X>{\centering\arraybackslash}X}
    \toprule
    Domain & Rotational speed \\
    \midrule
    $B_1$    & 600 r/min  \\
    $B_2$    & 800 r/min  \\
    $B_3$    & 1000 r/min  \\
    \bottomrule
    \end{tabularx}%
  \label{tab:JNU}%
\end{table}%

\subsection{Baselines}
The comparative analysis includes a range of transfer learning and domain adaptation strategies, categorized as follows:

\noindent\textbf{Metric-based UDA Methods} aim to reduce distributional discrepancies between source and target domains. Joint Adaptation Network (JAN) \cite{long2017deep} and Correlation Alignment (CORAL) \cite{sun2016deep}, all of which focus on domain alignment.

\noindent\textbf{Adversarial-based UDA Methods} employ adversarial training to achieve domain adaptation and demonstrate strong generalization capabilities. The evaluated algorithms are Domain Adversarial Neural Network (DANN)\cite{ganin2016domain} and Adversarial Discriminative Domain Adaptation (ADDA) \cite{Tzeng_2017_CVPR}.

\noindent\textbf{SFDA Methods} operate without accessing source domain data. The selected algorithms include Source Hypothesis Transfer (SHOT) \cite{liang2020we}, a pioneering SFDA method using pseudo-label cross-entropy and entropy regularization; Source-Free Adaptation Diagnosis (SFAD) \cite{jiao2022source}, which replaces entropy regularization with a kernel matrix loss; and Source-Free Cluster Adaptation (SF-CA) \cite{zhu2023source}, which applies R-drop techniques to improve generalization.

\begin{table*}[htbp]
  \centering
  \caption{Network Architecture Details Of The Proposed Model}
  \resizebox{\textwidth}{!}{%
    \begin{tabular}{>{\centering\arraybackslash}m{5em} >{\raggedright\arraybackslash}m{22em} >{\centering\arraybackslash}m{11em} >{\centering\arraybackslash}m{15em}}
    \toprule
    \textbf{Modules} & \multicolumn{1}{c}{\textbf{Layers}} & \textbf{Parameters} & \textbf{Output Shape} \\
    \midrule
    \multirow{20}[1]{=}{\centering\textbf{Feature Encoding}} 
      & Cov → BN 
      & Input channels: 1, Output channels: 64, Kernel size: 7, Stride: 2, Padding: 3 
      & [batch\_size, seq\_length / 2, 64] \\
      & \rule[-1ex]{0pt}{5ex}Conv → BN → ReLU → Conv → BN → ResAdd → ReLU → Conv → BN → ReLU → Conv → BN → ResAdd → ReLU 
      & Output channels: 64, Stride: 1, Padding: 1 
      & [batch\_size, seq\_length / 2, 64] \\
      & \rule[-1ex]{0pt}{5ex}Conv → BN → ReLU → Conv → BN → ResAdd → ReLU → Conv → BN → ReLU → Conv → BN → ResAdd → ReLU 
      & Output channels: 128, Stride: 2, Padding: 1 
      & [batch\_size, seq\_length / 4, 128] \\
      & \rule[-1ex]{0pt}{5ex}Conv → BN → ReLU → Conv → BN → ResAdd → ReLU → Conv → BN → ReLU → Conv → BN → ResAdd → ReLU 
      & Output channels: 256, Stride: 2, Padding: 1 
      & [batch\_size, seq\_length / 8, 256] \\
      & \rule[-1ex]{0pt}{5ex}Conv → BN → ReLU → Conv → BN → ResAdd → ReLU → Conv → BN → ReLU → Conv → BN → ResAdd → ReLU 
      & Output channels: 512, Stride: 2, Padding: 1 
      & [batch\_size, 512, seq\_length / 16] \\
      & \rule[-1ex]{0pt}{5ex}Average Pool → Dropout 
      & Output channels: 1, Dropout rate: 0.1 
      & [batch\_size, 1, 512] \\
      & \rule[-1ex]{0pt}{5ex}FC → BN 
      & 
      & [batch\_size, 256, 1] \\
    \midrule
    \textbf{Classification} 
      & FC (with wn) → Softmax 
      & 
      & [batch\_size, class\_number] \\
    \bottomrule
    \end{tabular}%
  }
  \label{tab:model}
\end{table*}

\subsection{Implementation Details}

The proposed SDALR method was implemented using the PyTorch framework. To ensure fair comparisons, the baseline UDA methods were implemented using the Transfer-Learning-Library, while SHOT, SFAD and SF-CA were independently reproduced.

\textbf{Network Architecture}: This study adopts ResNet-18 as the backbone network, modified specifically for one-dimensional input to accommodate the characteristics of vibration signal data. The network consists of two main modules: a feature extraction module and a classification module. The feature extraction module includes the modified backbone and the intermediate structure, which collectively facilitate hierarchical feature extraction from the input signals. The classification module incorporates a fully connected layer with weight normalization (FC with wn), a technique that re-parameterizes weights into magnitude and direction. This design improves numerical stability during training and accelerates convergence. Detailed network parameters are summarized in Table \ref{tab:model}, where Conv represents one-dimensional convolution, BN denotes batch normalization, ReLU indicates the rectified linear unit activation function, ResAdd refers to residual addition and FC stands for fully connected layers. 

\textbf{Experimental Setup}: Training employed the stochastic gradient descent (SGD) optimizer with a batch size of 64. For source model training, the initial learning rate was set to  $7\times10^{-3}$, with training spanning 10 epochs. For target model training, a lower learning rate ($5\times10^{-4}$ for the PU and JNU dataset) was used to prevent overfitting, extending training to 20 epochs. A dynamic learning rate decay mechanism was implemented as:
	$\mathrm{lr}=\mathrm{lr}_0\cdot(1+10p)^{0.75}$
    where $p$ represents the training progress linearly scaled from 0 to 1 and $lr_0$ is the initial learning rate.

\subsection{Results}

\begin{table*}[th]
  \centering
  \caption{Diagnosis Results (\%) on Dataset PU}
  \resizebox{1\textwidth}{!}{%
  \tiny
    \begin{tabular}{ccccccccc}
    \toprule
    Methods & SF    & $A_1$→$A_2$ & $A_1$→$A_3$ & $A_2$→$A_1$ & $A_2$→$A_3$ & $A_3$→$A_1$ & $A_3$→$A_2$ & Average \\
    \midrule
    JAN\cite{long2017deep} & \ding{55}     & 87.68 & 92.95 & 78.67 & 80.28 & 90.97 & 86.8  & 86.23 \\
    CORAL\cite{sun2016deep} & \ding{55}     & 81.22 & 91.88 & 76.81 & 77.13 & 90.26 & 80.02 & 82.89 \\
    DANN\cite{ganin2016domain} & \ding{55}     & 89.69 & 95.8  & 83.64 & 86.07 & 94.8  & 90.94 & 90.16 \\
    ADDA\cite{Tzeng_2017_CVPR} & \ding{55}     & 82.73 & 93.7  & 75.61 & 75.4  & 91.7  & 80.77 & 83.32 \\
    SHOT\cite{liang2020we} & \ding{51}     & 86.41 & 94.58 & 82.58 & 85.12 & 92.95 & 87.88 & 88.25 \\
    SFAD\cite{jiao2022source} & \ding{51}     & 85.33 & 93.58 & 80.36 & 86.23 & 90.96 & 86.9  & 87.23 \\
    SF-CA\cite{zhu2023source} & \ding{51}     & 89.61 & 96.7  & 86.47 & 86.27 & 95.03 & 90.71 & 90.8 \\
    LEAD\cite{} & \ding{51}     & \textbf{95.6} & 98.4  & 84.3 & 84.9 & 98.9 & 84.5? & 91.1 \\
    SDALR & \ding{51}     & 87.03 & \textbf{99.95} & \textbf{96.19} & \textbf{99.73} & \textbf{99.96} & \textbf{97.84} & \textbf{96.78} \\
    \bottomrule
    \end{tabular}%
  }
  \label{tab:result_pu}
\end{table*}%

\begin{table*}[th]
  \centering
  \caption{Diagnosis Results (\%) on Dataset JNU}
  \resizebox{\textwidth}{!}{%
  \tiny
    \begin{tabular}{ccccccccc}
    \toprule
    Methods & SF    & $B_1$→$B_2$ & $B_1$→$B_3$ & $B_2$→$B_1$ & $B_2$→$B_3$ & $B_3$→$B_1$ & $B_3$→$B_2$ & Average \\
    \midrule
    JAN\cite{long2017deep} & \ding{55}     & 96.28 & 96.01 & 95.1  & 97.79 & 91.79 & 97.57 & 95.76 \\
    CORAL\cite{sun2016deep} & \ding{55}     & 94.37 & 92.66 & 92.75 & 96.25 & 89.3  & 94.73 & 93.34 \\
    DANN\cite{ganin2016domain} & \ding{55}     & 98.46 & 97.78 & 95.68 & 98.37 & 93.29 & 98.68 & 97.04 \\
    ADDA\cite{Tzeng_2017_CVPR} & \ding{55}     & 95.7  & 94.64 & 85.32 & 91.37 & 89.72 & 95.37 & 92.02 \\
    SHOT\cite{liang2020we} & \ding{51}     & 94.31 & 88.35 & 90.27 & 84.37 & 92.05 & 97.29 & 91.11 \\
    SFAD\cite{jiao2022source} & \ding{51}     & 99.14 & 98.01 & 94.59 & 98.43 & 93.41 & 99.16 & 97.12 \\
    SF-CA\cite{zhu2023source} & \ding{51}     & 99.48 & 98.59 & 94.71 & 98.81 & 93.28 & 99.6  & 97.41 \\
    LEAD\cite{} & \ding{51}     & 96.1 & 96.1 & 92.0 & 95.4 & 87.0 & 91.8  & 93.0 \\
    SDALR & \ding{51}     & \textbf{99.98} & \textbf{100} & \textbf{96.41} & \textbf{100} & \textbf{94.71} & \textbf{99.92} & \textbf{98.50} \\
    \bottomrule
    \end{tabular}%
  }
  \label{tab:result_jnu}
\end{table*}%

The proposed SDALR method was evaluated against baselines on the PU and JNU datasets, with results summarized in Tables \ref{tab:result_pu} and \ref{tab:result_jnu}. In these tables, "SF" denotes the source-free setting and "$A_1$→$A_2$" indicates domain adaptation from domain $A_1$ to $A_2$. Bolded results indicate the best performance for each task.

\textbf{Comparison with Traditional UDA Methods}: Unlike UDA methods, which require source domain data and raise data privacy concerns, SDALR achieves comparable or superior performance without accessing labeled source data. On the PU dataset, SDALR improves average accuracy by 6.52\% over the best UDA method, DANN, and by 1.46\% on the JNU dataset.

\textbf{Comparison with SFDA Methods}: Among SFDA methods, SHOT has demonstrated potential in image classification, while SFAD targets multi-feature domain bearing diagnosis. SF-CA enhances target adaptation using conditional autoencoders. However, SDALR consistently outperforms these methods, achieving 5.98\% higher accuracy than SF-CA on the PU dataset and 1.09\% on the JNU dataset, highlighting its superior adaptability and diagnostic precision. 

\subsection{Analysis}

\subsubsection{Classification Performance Analysis via Confusion Matrix}
\begin{figure*}[!th]
  \centering
  \includegraphics[width=1\linewidth]{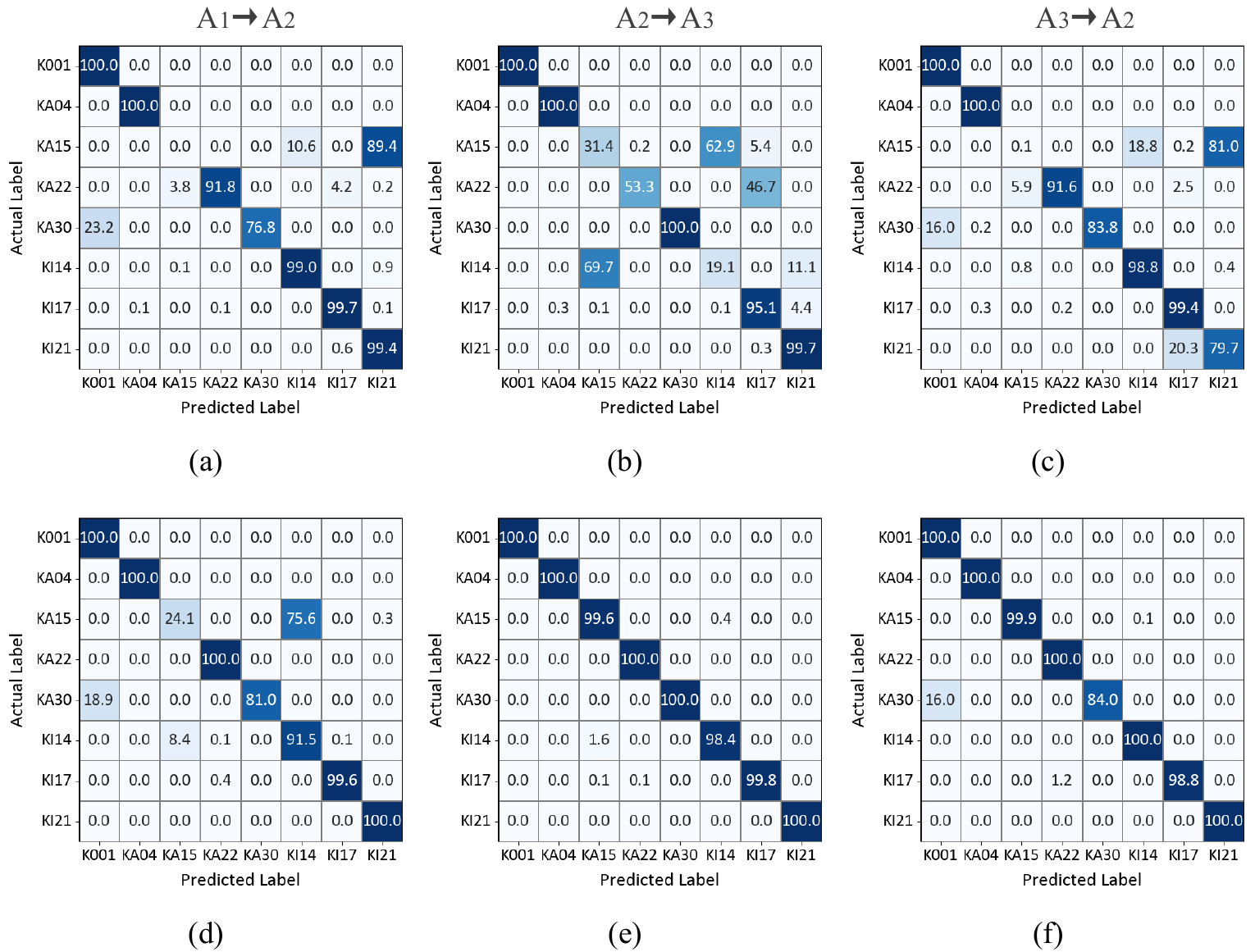}
  \caption{Confusion matrix comparison of SF-CA and SDALR on PU datasets to evaluate classification performance across different tasks. Subplot (a)(b)(c) represent the confusion matrices of the SF-CA method, while subplot (d)(e)(f) correspond to the confusion matrices of the SDALR method}
  \label{fig:cm_pu}
\end{figure*}

\begin{figure*}[!th]
  \centering
  \includegraphics[width=0.8\linewidth]{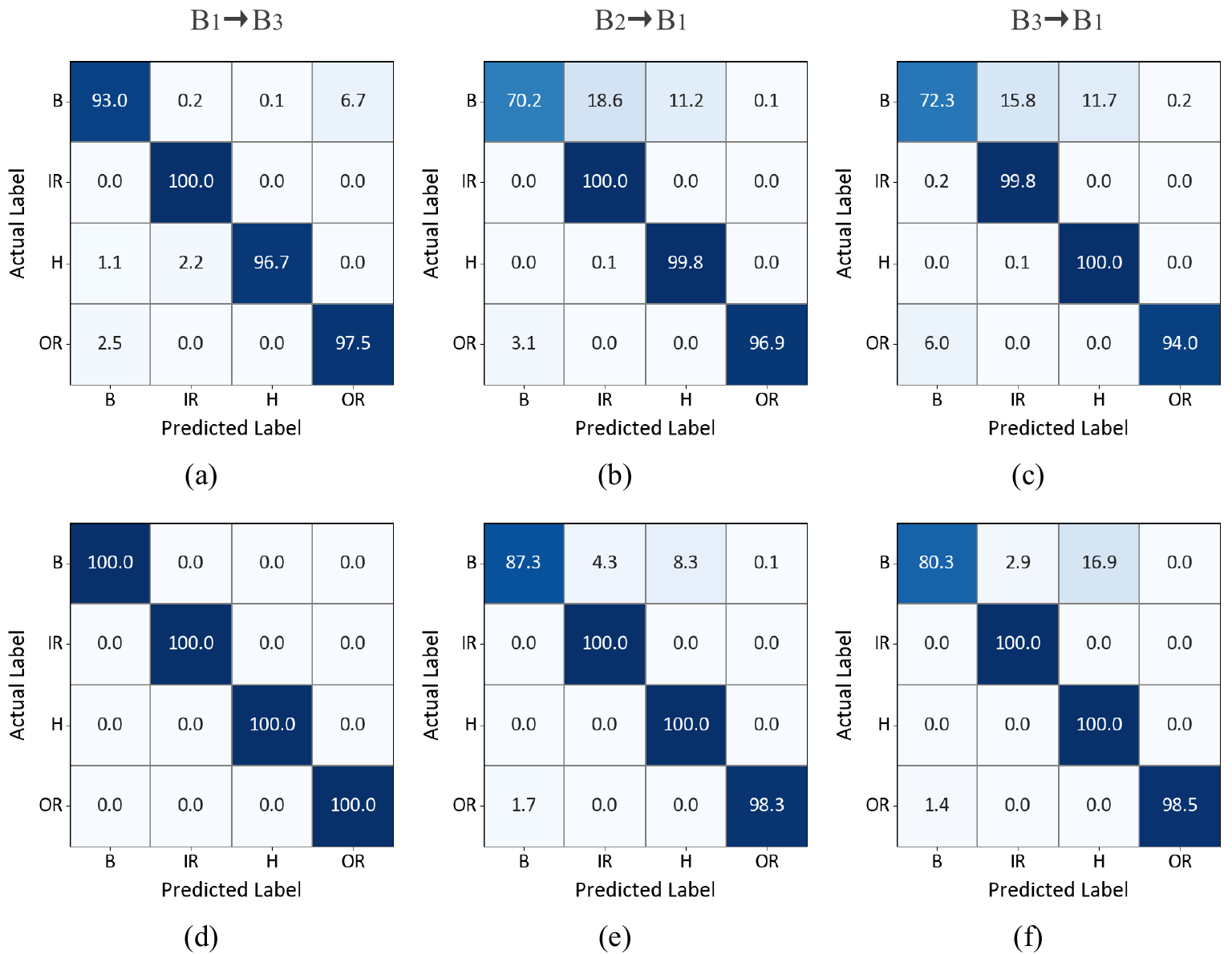}
  \caption{Confusion matrix comparison of SF-CA and SDALR on JNU datasets to evaluate classification performance across different tasks. Subplot (a)(b)(c) represent the confusion matrices of the SF-CA method, while subplot (d)(e)(f) correspond to the confusion matrices of the SDALR method}
  \label{fig:cm_jnu}
\end{figure*}

To gain deeper qualitative insights, feature representations were visualized using t-SNE, focusing on the performance of SDALR and the state-of-the-art model SF-CA on two datasets. Tasks $A_1$→$A_2$, $A_2$→$A_3$ and $A_3$→$A_1$ were used for the PU dataset, while tasks $B_1$→$B_3$, $B_2$→$B_3$ and $B_3$→$B_1$ were analyzed for the JNU dataset. Figures \ref{fig:visual_pu} and \ref{fig:visual_jnu} display the feature representations extracted from the final layer of the feature encoder, where different colors correspond to different categories. For the PU dataset, the visualizations cover eight bearing states, while for the JNU dataset, four bearing states are depicted. Unlike traditional unsupervised domain adaptation methods that align features between source and target domains, this study focuses solely on target-domain features due to the source-free adaptation setup. The results show that SDALR learns distinct and well-separated feature representations for each category, enabling precise fault diagnostics. In contrast, SF-CA exhibits less clear class separation.

\subsubsection{Hyperparameter Tuning}
\begin{figure}[!th]
  \centering
  \includegraphics[width=1\linewidth]{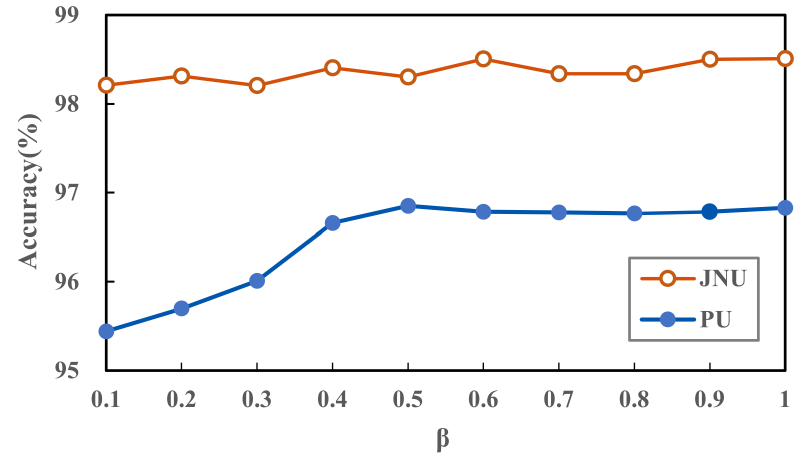}
  \caption{Bearing fault diagnosis accuracy curves for hyper-parameter $\beta$.}
  \label{fig:K}
\end{figure}

\begin{figure}[!th]
  \centering
  \includegraphics[width=1\linewidth]{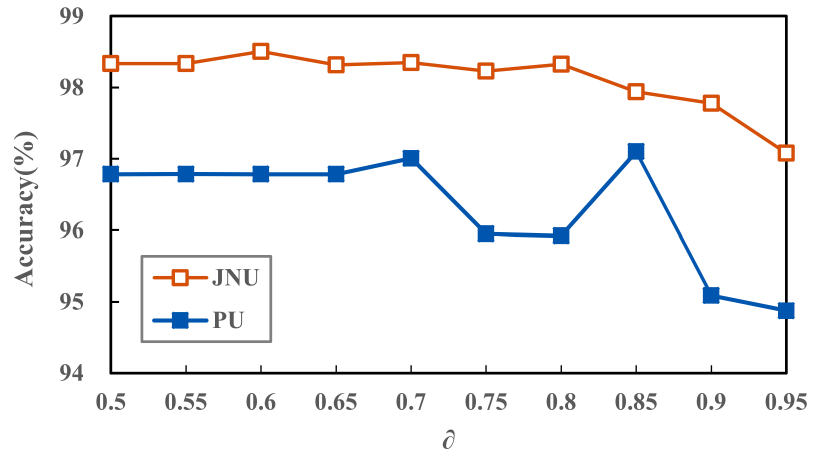}
  \caption{Bearing fault diagnosis accuracy curves for hyper-parameter $\partial$.}
  \label{fig:threshold}
\end{figure}

To assess the impact of hyperparameter selection on the performance of SDALR, experiments were conducted using the PU and JNU datasets, covering all six tasks in each dataset. The analysis focused on two critical hyperparameters: the $\beta$ value and the $\partial$ value. The $\beta$ value determines the weighting of the repulsion component in the $L_{car}$ loss, balancing the aggregation of similar samples with the dispersion of dissimilar samples. The $\partial$ specifies the similarity criterion used in the pseudo-label voting strategy. During the experiments, the $\partial$ was initially fixed at 0.6 and the $\beta$ was varied from 0.1 to 1.0 in increments of 0.1. Subsequently, $\beta$ was fixed at 0.6 and the $\partial$ was adjusted from 0.5 to 0.95 in increments of 0.05. Figures \ref{fig:K} depict the accuracy trends corresponding to varying $\beta$ for the PU and JNU datasets, respectively, while Figures \ref{fig:threshold} illustrate the accuracy variations under different thresholds for both datasets.

The results reveal that, with the threshold fixed at 0.6, the accuracy trends with varying $\beta$ differ somewhat between the two datasets but share a general pattern: accuracy exhibits an overall upward trend, albeit with fluctuations, as $\beta$ increases. To address the imbalance in the number of similar and dissimilar samples, adjustments to the hyperparameters were explored to balance their weights. Experimental findings indicate that increasing the weight of the repulsion component for dissimilar samples positively impacts accuracy, underscoring the importance of maintaining clear inter-class distinctions in feature space to enhance model performance.

When the $\beta$ is fixed at 0.6, threshold $\partial$ experiments exhibit a consistent trend across both datasets, as shown in the Figure \ref{fig:threshold}. As the threshold increases from 0.5 to 0.95, accuracy initially remains stable but eventually declines. This decline suggests that excessively strict thresholds (e.g., 0.95) overly constrain the selection range of samples, thereby hindering model performance. Conversely, as the threshold becomes less restrictive, accuracy stabilizes, indicating that the current sample selection mechanism effectively mitigates the adverse effects of unreliable samples while preserving robust model performance across various threshold settings.

\subsubsection{Ablation Studies}
Our model is optimized using multiple mechanisms, including cohesion and repulsion loss ($L_{car}$), pseudo-label voting (voting) and entropy maximization loss ($L_{uem}$). To evaluate the effectiveness of each module, we conducted an ablation study. In the initial experiment, cohesion and repulsion loss ($L_{car}$), pseudo-label voting (voting), and entropy maximization loss ($L_{uem}$) were removed, leaving only label-smoothing cross-Entropy loss ($L_{lsc}$) and information maximization loss ($L_{im}$) as the baseline method. This configuration is analogous to the SHOT method. Subsequently, the modules were added incrementally to assess their contributions to improving model performance.

The detailed results of the ablation study on two datasets are presented in Table \ref{tab:Ablation_PU} and Table \ref{tab:Ablation_JNU}. Comparative analysis indicates that the model's diagnostic performance improves with the inclusion of each additional module. These findings highlight the positive contributions of each component to the overall performance, validating the proposed approach and supporting the rationality of its modular design.

\begin{table*}[htbp]
  \centering
  \caption{Results(\%) of Ablation Studies in PU}
  \resizebox{\textwidth}{!}{%
  \footnotesize
    \begin{tabular}{l|ccccccc}
    \toprule
    \multicolumn{1}{c|}{Methods} & $A_1$→$A_2$ & $A_1$→$A_3$ & $A_2$→$A_1$ & $A_2$→$A_3$ & $A_3$→$A_1$ & $A_3$→$A_2$ & Average \\
    \midrule
    $L_{lsc}$ + $L_{im}$ & 86.41  & 94.58  & 82.58  & 85.12  & 92.95  & 87.88  & 88.25  \\
    $L_{lsc}$ + $L_{im}$ + $L_{car}$ & 87.05  & 99.99  & 87.38  & 99.63  & 99.97  & 97.51  & 95.26  \\
    $L_{lsc}$ + $L_{im}$ + $L_{car}$ + Voting & 87.62  & 99.98  & 87.88  & 99.89  & 99.97  & 97.68  & 95.50  \\
    $L_{lsc}$ + $L_{im}$ + $L_{car}$ + Voting + $L_{uem}$ & 87.03  & 99.95  & 96.19  & 99.73  & 99.96  & 97.84  & 96.78  \\
    \bottomrule
    \end{tabular}%
  }
  \label{tab:Ablation_PU}
\end{table*}%

\begin{table*}[htbp]
  \centering
  \caption{Results(\%) of Ablation Studies in JNU}
  \resizebox{\textwidth}{!}{%
  \footnotesize
    \begin{tabular}{l|ccccccc}
    \toprule
    \multicolumn{1}{c|}{Methods} & $B_1$→$B_2$ & $B_1$→$B_3$ & $B_2$→$B_1$ & $B_2$→$B_3$ & $B_3$→$B_1$ & $B_3$→$B_2$ & Average \\
    \midrule
    $L_{lsc}$ + $L_{im}$ & 94.31  & 88.35  & 90.27  & 84.37  & 92.05  & 97.29  & 91.11  \\
    $L_{lsc}$ + $L_{im}$ + $L_{car}$ & 99.98  & 100.00  & 96.04  & 99.99  & 93.66  & 99.94  & 98.27  \\
    $L_{lsc}$ + $L_{im}$ + $L_{car}$ + Voting & 99.96  & 100.00  & 96.34  & 99.99  & 94.49  & 99.95  & 98.46  \\
    $L_{lsc}$ + $L_{im}$ + $L_{car}$ + Voting + $L_{uem}$ & 99.98  & 100.00  & 96.41  & 100.00  & 94.71  & 99.92  & 98.50  \\
    \bottomrule
    \end{tabular}%
  }
  \label{tab:Ablation_JNU}%
\end{table*}%

\subsubsection{Visualization}
\begin{figure}[!h]
  \centering
  \includegraphics[width=1\linewidth]{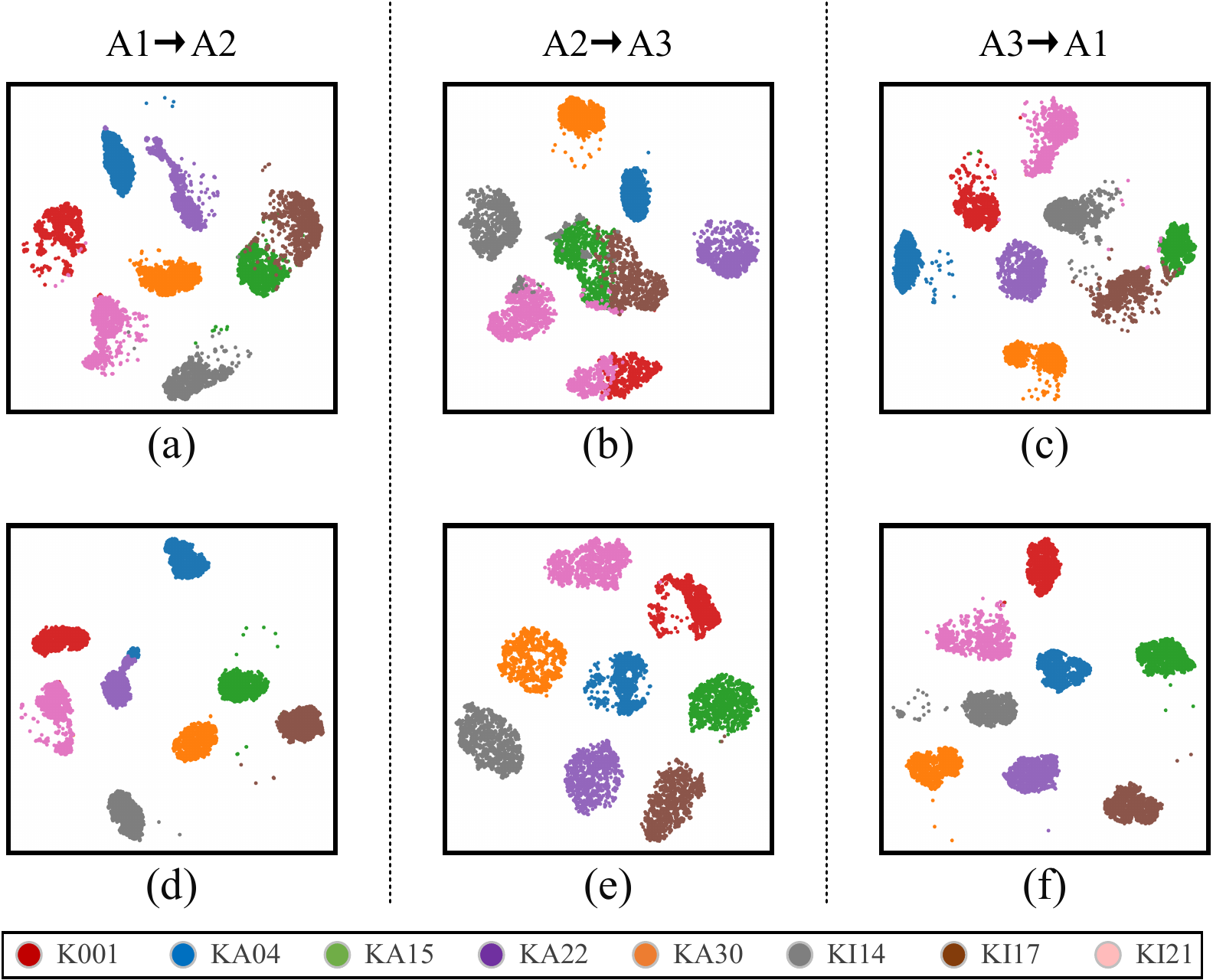}
  \caption{Qualitative visualization using t-SNE to illustrate the distribution patterns across PU datasets from different domains. Subplot (a)(b)(c) represent the visualization results of SF-CA, while subplot (d)(e)(f) represent the visualization results of SDALR.}
  \label{fig:visual_pu}
\end{figure}

\begin{figure}[!h]
  \centering
  \includegraphics[width=1\linewidth]{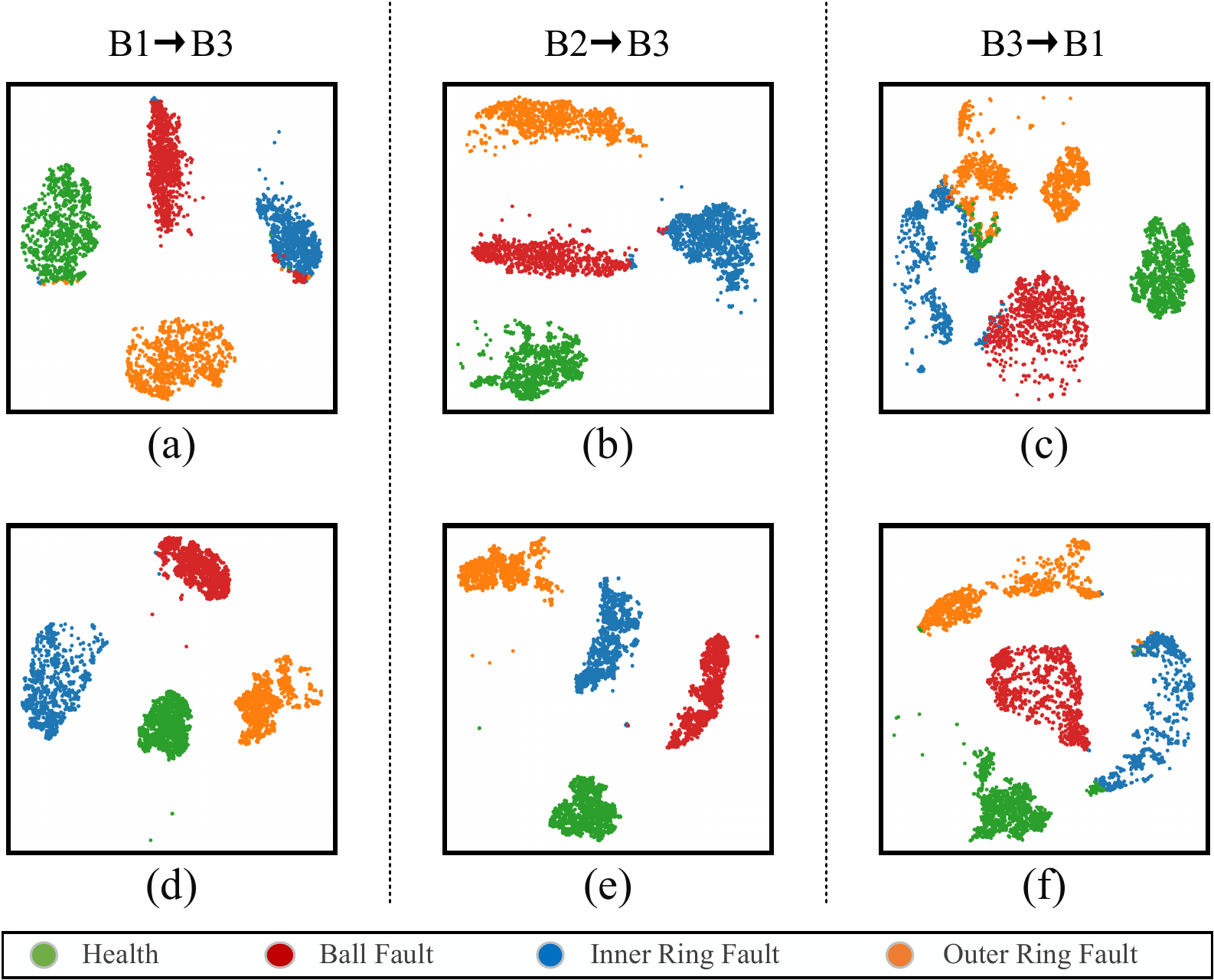}
  \caption{Qualitative visualization using t-SNE to illustrate the distribution patterns across PU datasets from different domains. Subplot (a)(b)(c) represent the visualization results of SF-CA, while subplot (d)(e)(f) represent the visualization results of SDALR.}
  \label{fig:visual_jnu}
\end{figure}

To gain deeper qualitative insights, feature representations were visualized using t-SNE, focusing on the performance of SDALR and the state-of-the-art model SF-CA on two datasets. Tasks $A_1$→$A_2$, $A_2$→$A_3$, $A_3$→$A_1$, $B_1$→$B_3$, $B_2$→$B_3$, $B_3$→$B_1$ were selected for evaluation. Figures \ref{fig:visual_pu} and \ref{fig:visual_jnu} display the feature representations extracted from the final layer of the feature encoder, where different colors correspond to different categories. For the PU dataset, the visualizations cover eight bearing states, while for the JNU dataset, four bearing states are depicted. Unlike traditional unsupervised domain adaptation methods that align features between source and target domains, this study focuses solely on target-domain features due to the source-free adaptation setup. The results show that SDALR learns distinct and well-separated feature representations for each category, enabling precise fault diagnostics. In contrast, SF-CA exhibits less clear class separation. These findings confirm that SDALR achieves superior feature learning, further validating its effectiveness over SF-CA.

\section{Conclusion}\label{sec-con}
This paper proposes an innovative SFDA method to address three core challenges in source-free cross-domain bearing fault diagnosis: ensuring discriminability, and diversity in the feature space, improving pseudo-label quality and reducing negative transfer from unreliable target samples. The approach integrates pseudo-label-guided clustering for feature structuring, a data-augmentation-based voting strategy for reliable pseudo-labels, and entropy maximization to minimize the impact of unreliable samples. Experimental results show that the method outperforms state-of-the-art methods in source-free cross-domain fault diagnosis.

While effective, the method's success heavily depends on the quality of the source model. Future work should enhance source model generalization, leverage reliable information to improve adaptability, and extend this approach to dynamic adaptation scenarios and complex industrial applications for broader relevance.

\section*{Declaration of competing interest}

The authors declare that they have no known competing financial interests or personal relationships that could have appeared to influence the work reported in this paper.

\section*{Acknowledgments}

Preparation of this review article was supported in part by the National Natural Science Foundation of China (No.62176069), in part by the Natural Science Foundation of Guangdong Province (No. 2019A1515011076), in part by the Innovation Group of Guangdong Education Department, China (No. 2020KCXTD014), in part by the Special Program of Education Bureau of Guangdong Province, China (No. 2024ZDZX1035), in part by the Open Project of Key Laboratory of Data Science and Intelligence Application of Fujian Province, Chian (No. D202002), and in part by Fujian Provincial Natural Science Foundation of China (No.2022J01913).

\bibliography{KBS-manu-2024v1}

\end{document}